\documentclass[preprint,12pt]{elsarticle}

\usepackage{amssymb}
\usepackage{subfigure}
\usepackage{setspace}
\usepackage{geometry} 
\usepackage{xcolor}

\usepackage{multirow}
\usepackage{amsmath}
\usepackage{mathrsfs}
\usepackage{algorithm}
\usepackage{algpseudocode}

\usepackage{lineno}

\begin{document}

\begin{frontmatter}




\title{URSimulator: Human-Perception-Driven Prompt Tuning for Enhanced Virtual Urban Renewal via Diffusion Models}


\author[inst1]{Chuanbo Hu} 
\affiliation[inst1]{organization={University at Albany, State University of New York},
            city={Albany},
            state={New York},
            country={United States}}

\author[inst2]{Shan Jia}
\affiliation[inst2]{organization={Google},
            city={Mountain View},
            state={California},
            country={United States}}

\author[inst1]{Xin Li}

\begin{abstract}

Tackling Urban Physical Disorder (e.g., abandoned buildings, litter, messy vegetation, and graffiti) is essential, as it negatively impacts the safety, well-being, and psychological state of communities. Urban Renewal is the process of revitalizing these neglected and decayed areas within a city to improve their physical environment, and quality of life for residents. Effective urban renewal efforts can transform these environments, enhancing their appeal and livability. However, current research lacks simulation tools that can quantitatively assess and visualize the impacts of urban renewal efforts, often relying on subjective judgments. Such simulation tools are essential for planning and implementing effective renewal strategies by providing a clear visualization of potential changes and their impacts. This paper presents a novel framework that addresses this gap by using human perception feedback to simulate the enhancement of street environment. We develop a prompt tuning approach that integrates text-driven Stable Diffusion with human perception feedback. This method iteratively edits local areas of street view images, aligning them more closely with human perceptions of beauty, liveliness, and safety. Our experiments show that this framework significantly improves people's perceptions of urban environments, with increases of 17.60\% in safety, 31.15\% in beauty, and 28.82\% in liveliness. In comparison, other advanced text-driven image editing methods like DiffEdit only achieve improvements of 2.31\% in safety, 11.87\% in beauty, and 15.84\% in liveliness. We applied this framework across various virtual scenarios, including neighborhood improvement, building redevelopment, green space expansion, and community garden creation. The results demonstrate its effectiveness in simulating urban renewal, offering valuable insights for real-world urban planning and policy-making. This method not only enhances the visual appeal of neglected urban areas but also serves as a powerful tool for city planners and policymakers, ultimately improving urban landscapes and the quality of life for residents.

\end{abstract}



\begin{highlights}
    
    \item \textbf{AI-Driven Urban Renewal Advancements:} We showcase how cutting-edge AI techniques can simulate enhancements in the beauty, safety, and liveness of urban environments.
    
    \item \textbf{Innovative Human Perception Integration:} This study introduces a new simulation approach to urban renewal by incorporating human feedback into text-driven Stable Diffusion, enhancing the effectiveness of urban renewal.

    \item \textbf{Proven Framework Effectiveness:} Our experiments reveal at least a 17\% enhancement in human perception scores for urban renewal simulations, surpassing the performance of existing text-driven image editing models.

    \item \textbf{Adaptable Urban Applications:} The successful application of the framework in four different urban scenarios demonstrates its adaptability and effectiveness in real-world situations.

    \item \textbf{Strategic Urban Planning Tool:} This AI-driven framework provides city planners and policymakers with a powerful simulation tool to transform urban spaces into more livable and appealing environments.
    
\end{highlights}

\begin{keyword}
  Urban renewal \sep Urban physical disorder \sep Street view imagery \sep Prompt engineering \sep Stable diffusion \sep Human perception
\end{keyword}

\end{frontmatter}


\section{Introduction}\label{sec:1}


Urban landscapes are dynamic environments that significantly influence the quality of life of their inhabitants. As cities expand and infrastructure ages, signs of Urban Physical Disorder (UPD) often emerge \cite{sampson1999systematic}. UPD refers to visible decay and deterioration in urban settings, manifesting as abandoned buildings, litter, graffiti, refuse accumulation, vandalized properties, pothole-ridden roads, and overgrown public spaces \cite{jones2011eyes}. These signs of neglect not only decrease resident satisfaction but also contribute to a range of mental health issues, including anxiety, loneliness, and depression \cite{bjornstrom2013social,fathi2020role,dassopoulos2012neighborhood,sprott2009effect,kang2020review,sampson2001disorder}. Addressing UPD is essential, as its presence is closely linked to increased crime rates and social disturbances, making it a critical issue for public health and urban well-being.

Urban renewal offers a powerful solution to these challenges by transforming neglected areas into well-maintained, vibrant spaces.
Effective urban renewal can significantly enhance human perceptions of safety, liveliness, and overall quality of life. For instance, introducing green spaces, recreational areas, and aesthetically pleasing architectural designs has been shown to reduce crime and improve residents' sense of security \cite{benkHo2016crime,ng2018urban}. Projects like the High Line in New York City, which repurposed an abandoned railway into a thriving public park, have led to increased local economic activity and stronger community cohesion \cite{lindsey2001access}. Despite these benefits, the significant costs associated with urban renewal projects necessitate using effective simulation tools to model and plan improvements \cite{accordino2000addressing}. 

Traditional visual simulation tools such as CityEngine \cite{CityEngine2020}, UrbanSim \cite{Waddell2002}, and ArcGIS Urban \cite{ArcGISUrban2020} assist urban planners in visualizing potential changes. However, these tools often fall short of capturing the quantitative impacts of urban renewal, particularly on human perceptions such as safety, beauty, and liveliness. This gap highlights the need for more advanced and responsive tools to simulate urban renewal's dynamic changes better.
Recent advancements in technology, particularly the use of Street View Imagery (SVI), have provided urban planners with a detailed and virtually accessible dataset that mirrors real-world environments \cite{anguelov2010google,rundle2011using,gebru2017using,li2015assessing,hu2023saliency}. SVI has been effectively used to assess human perceptions of urban spaces, offering valuable insights into street-level aesthetics and safety \cite{zhang2018measuring,larkin2021predicting,wang2019perceptions}. However, existing methods primarily focus on static analysis, limiting their ability to simulate the dynamic changes required for effective urban renewal. This limitation underscores the importance of developing more interactive and adaptable urban planning strategies to address the evolving needs of urban communities.

\begin{figure}[!h]
 \centering
 \includegraphics[width=1\linewidth]{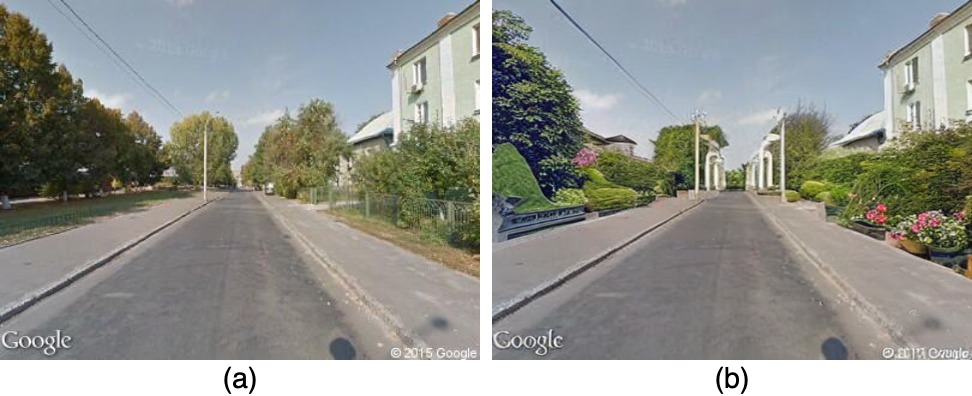}
 \caption{Example of Urban Renewal Simulation Using Street View Images: (a) Street view depicting physical disorder factor (messy vegetation); (b) Vegetation renewal achieved through image editing.}
  \label{fig:example}
\end{figure}

Generative AI presents a promising solution to this challenge, particularly in image manipulation and editing. Studies have demonstrated the potential of AI-driven image editing to modify urban scenes by adding elements such as trees, benches, or entire buildings to street view images \cite{hong2018learning,ntavelis2020sesame}. These capabilities allow for the simulation of various urban renewal scenarios, offering a flexible and interactive tool for urban planners.  For example, Figure \ref{fig:example} (a) shows an original street view image, while Figure \ref{fig:example} (b) illustrates the same view after applying image editing to transform the vegetation into a well-maintained garden. The improvements in the street landscape are noticeable, providing a visual representation of the potential effects of urban renewal before implementation. This dynamic approach lets stakeholders explore multiple design options and make informed, data-driven decisions.
Inspired by the potential of Generative AI, our framework leverages these advanced techniques to enhance urban renewal simulations. Our work makes the following key contributions:

\begin{itemize}
    \item \textbf{Introduction of a Framework Using SOTA Stable Diffusion}: We present a novel framework that employs the state-of-the-art (SOTA) stable diffusion method, allowing for the simulation of urban renewal in a virtual environment. This technique provides a realistic representation of potential urban modifications, offering a glimpse into the future of renewed urban spaces.
    
    \item \textbf{Human Perception-Guided Prompt Engineering}: By integrating human psychology into our approach, we pioneer the concept of human perception-guided prompt engineering. This method ensures that urban renewal strategies are technically feasible and resonate with the psychological needs and preferences of city residents, building a brighter, more harmonious urban future.
    
    \item \textbf{A Comprehensive Tool for Urban Planners}: Our research culminates in developing a robust tool tailored for urban planners. This tool, grounded in our innovative framework and prompt engineering approach, provides a tangible means to evaluate, visualize, and implement urban renewal strategies, making the planning process more efficient and impactful.
    
\end{itemize}

These contributions collectively advance the field of urban renewal by integrating advanced AI techniques with human-centered design principles.

The remainder of this paper is organized as follows: In Section \ref{sec:2}, we provide an extensive review of related work, situating our research within the broader context of urban renewal and technological interventions. Section \ref{sec:3} delves into the detailed methodology we employed, elucidating the intricacies of our approach. This is followed by Section \ref{sec:4}, where we present the experiments conducted and their corresponding results. In Section \ref{sec:5}, we engage in comprehensive discussions, interpreting our findings and drawing connections with existing literature. Finally, Section \ref{sec:6} concludes the paper, summarizing our contributions and hinting at potential future directions in this domain.

\section{Related Work}\label{sec:2}

Urban renewal and its psychological implications have garnered interest from urban planners, psychologists, and technologists. As technology evolves, so do the methods and strategies for addressing urban challenges. This section reviews key literature that has informed our approach, grouped into four primary themes.

\subsection{Urban Physical Disorder and Urban Renewal}

The relationship between UPD and the mental well-being of city inhabitants has been extensively studied. Visible signs of disorder, such as graffiti, litter, and abandoned buildings, are consistently associated with increased psychological distress among urban residents. These signs of neglect contribute to feelings of insecurity, lower quality of life, and elevated stress levels \cite{latkin2003stressful,ross2001neighborhood}.

Urban renewal, encompassing strategies to rejuvenate and revitalize urban areas, has often been proposed as a solution to mitigate these adverse effects. Historically, urban renewal efforts have focused on infrastructural and aesthetic improvements, such as repairing buildings, enhancing public spaces, and removing debris. For instance, the Broken Windows Theory posits that maintaining and monitoring urban environments to prevent small crimes like vandalism can foster an atmosphere of order and lawfulness, thus preventing more serious crimes \cite{wilson2017police}.

Recent studies emphasize the importance of incorporating psychological well-being into urban renewal strategies. Beyond physical improvements, urban renewal should create environments that enhance residents' mental health and overall quality of life. This includes designing spaces that promote social interaction, provide access to green areas, and ensure safety \cite{jennings2019relationship}. Research has shown that integrating green spaces and recreational areas into urban settings can significantly reduce stress and improve mental health outcomes \cite{taylor2006contact,de2003natural}.

For example, Kuo and Sullivan found that urban green spaces are associated with lower crime levels and improved mental health among residents \cite{kuo2001environment}. Similarly, Branas et al. demonstrated that greening vacant lots in urban neighborhoods led to reductions in gun violence and enhanced perceptions of safety \cite{branas2011difference}. These findings highlight the need for urban renewal projects to consider psychological well-being as a key component, aiming to create more sustainable and livable urban environments.

\subsection{Advanced Simulation Tools for Enhancing Urban Renewal Planning}

Advanced simulation tools in urban planning have revolutionized how planners envision, modify, and implement urban renewal projects. These tools enable precise visualizations and detailed scenario analyses, aiding in design and decision-making for urban development:

\begin{itemize}
    \item CityEngine: A software by Esri that excels in creating detailed 3D urban models, allowing planners to visualize the impacts of their renewal initiatives in real-time. It supports large-scale city models, essential for assessing proposed changes' visual and spatial effects \cite{CityEngine2020}.
    
    \item UrbanSim: A comprehensive modeling tool that simulates the interplay between land use, transportation, and housing policies over time. It helps urban planners predict how cities might evolve under various policy and investment scenarios, making it invaluable for strategic urban development \cite{Waddell2002}.
    
    \item ArcGIS Urban: Another tool from Esri, ArcGIS Urban combines spatial analytics with 3D visualization capabilities. It facilitates the evaluation of zoning, land use, and building density projects, supporting more informed urban planning decisions \cite{ArcGISUrban2020}.
    
\end{itemize}

While these tools provide sophisticated options for simulating and visualizing urban renewal projects, they have limitations. They often rely on static data, which may not reflect the dynamic nature of urban environments. Additionally, the integration of real-time human feedback is typically lacking, which is crucial for aligning simulations with the evolving needs and perceptions of urban residents. Furthermore, while these tools excel at visualizing physical changes, they do not adequately quantify qualitative aspects of urban renewal, such as perceived safety, beauty, and liveliness. Addressing these issues is essential for developing more interactive, responsive, and holistic urban planning solutions.

\subsection{Human Perception Based on Street View Imagery}

With the proliferation of digital platforms, SVI has become a valuable tool for urban researchers. It provides an immersive view of urban environments that enables large-scale analyses of human perceptions. Several studies have explored SVI's potential in assessing safety, aesthetics, and overall urban appeal.

Naik et al. developed a machine learning model that analyzed millions of street view images to generate a "StreetScore," which quantifies perceived safety based on visual cues \cite{naik2014streetscore}. Their findings underscored SVI's ability to effectively capture and quantify public perceptions of safety. Similarly, Quercia et al. examined the aesthetic qualities of urban environments using SVI, developing a model to assess factors like beauty, quietness, and happiness. Their research demonstrated that visual attributes significantly influence residents' perceptions of their surroundings \cite{quercia2014aesthetic}. Another study by Hara et al. combined crowdsourcing with SVI to identify street-level accessibility issues, revealing detailed insights into urban accessibility and infrastructure quality \cite{hara2013combining}.

The Place Pulse dataset 2.0 \cite{Min-MTDRAL-TIP2020}, introduced by Dubey et al., provides a large-scale repository of crowd-sourced human perceptions of urban environments, including ratings on safety, liveliness, and beauty. This dataset has been instrumental in studies aiming to quantify visual attributes of urban spaces and their impact on public perception. For example, Suel et al. used the Place Pulse dataset to analyze urban aesthetics and their relationship with human well-being, demonstrating the potential of combining SVI with crowd-sourced perception data to gain deeper insights into how urban environments affect human experiences \cite{roe2020urban}.

These studies collectively highlight SVI's potential as a powerful analytical tool. By leveraging this platform, researchers can extract valuable insights into how the public perceives urban environments, which can inform more effective urban renewal strategies.

\subsection{Image Editing and Prompt Engineering}

Text-driven image editing heavily relies on natural language, making these tools accessible and usable for a wide audience. Natural language enables users to specify detailed requirements without needing specialized knowledge, which is crucial for the broad adoption of AI technology in urban planning.

Notable models like Stable Diffusion generate high-quality images from text prompts, allowing for the creation of realistic urban scenarios based on user input \cite{rombach2021highresolution}. DiffEdit, another example, uses textual descriptions for precise image editing, enabling targeted modifications to specific areas of an image \cite{couairon2022diffedit}.

Despite advancements in multimodal large language models, such as ChatGPT-4Vision, which have improved applications in natural images, medical images, and app images \cite{yang2023dawn,wu2023can,jia2024can,hu2024multimodal}, they do not yet support localized image edits. This limitation underscores the importance of specialized image editing models like Stable Diffusion and DiffEdit, which can handle precise local edits.

The rise of text-driven AI models has also spurred the development of "prompt engineering" — crafting effective text prompts to achieve desired outcomes from AI systems \cite{white2023prompt,liu2022design}. In text-to-image generation, the specificity and clarity of language in a prompt can significantly influence the accuracy and quality of the generated images \cite{zhou2022large,hu2023unveiling}. As users and developers better harness the power of language in prompts, they can guide AI's creative processes more effectively, leading to improved and more tailored results.

In summary, the integration of advanced AI techniques, such as prompt engineering and text-driven image editing, offers urban planners new tools to visualize and implement urban renewal strategies more effectively. This will ultimately enhance urban environments and improve residents' quality of life.

\section{Methodology}\label{sec:3}

\subsection{Problem Formulation}

UPD detracts from the livability and attractiveness of urban environments by negatively impacting safety, aesthetics, and community well-being. To tackle the issue of urban physical disorder (UPD), this research introduces URSimulator, a comprehensive simulation framework designed to enhance human perception of urban renewal efforts. Given an input \emph{street view image} $SVI_{i} =[SVI_{1},SVI_{2}, ..., SVI_{n}]$, the simulation task involves several steps: 

\begin{itemize}
    \item UPD Detection: Detect whether $SVI_{i}$ indicates UPD. If UPD is detected, identify the specific \emph{factor} most closely associated with physical disorder $f_{i} =[f_{1},f_{2}, ..., f_{n}]$; 
    
    \item Targeted Interventions: Propose targeted interventions $I_{i} =[I_{1},I_{2}, ..., I_{n}]$. For each $I_{i}$, \emph{Textual prompts} $P$ guide the image editing  \( f_{\text{edit}}(P) \).
    
    \item Impact Evaluation: Evaluate the impact of these interventions $E_{i} =[E_{1},E_{2}, ..., E_{n}]$. Both the edited $SVI'_{i}$ and raw $SVI_{i}$ are evaluated using human perception metrics (\(M\)) to measure improvements in perceived safety, beauty, and liveliness. The intervention cycle continues until no noticeable improvement.
\end{itemize}

\subsection{Framework Overview}

\begin{figure}[!h]
 \centering
 \includegraphics[width=1\linewidth]{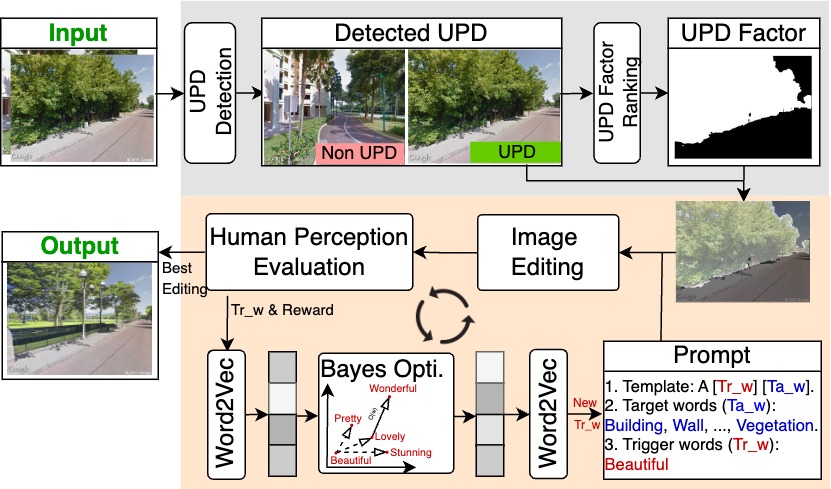}
 \caption{Framework of Prompt Tuning-Guided Urban Renewal. \textcolor{green}{Green} text represents input/output, \textcolor{red}{red} text indicates trigger words \(Tr_W\), and \textcolor{blue}{blue} text denotes target words \(Ta_W\) in the prompt; Gray Panel Represents UPDExplainer \cite{hu2023updexplainer}; Orange Panel Represents Proposed Framework }
  \label{fig:6}
\end{figure}

The proposed framework for urban renewal leverages advanced image editing techniques guided by textual prompts to address UPD. The framework comprises several key components that enhance human perception metrics, including safety, beauty, and liveliness in urban environments. The framework's structure is illustrated in Figure \ref{fig:6}.

The framework operates as follows:

\begin{itemize}
    \item \textbf{UPD Detection.} 
    The URSimulator begins by utilizing the UPDExplainer module $f_{u}(.)$ \cite{hu2023updexplainer}, which employs a binary classification network to determine whether the input image $SVI_i$ contains UPD.
    If UPD is detected, the module also identifies the factor most closely associated with the UPD $f_{i}$ and its location within the image.
    
    \item \textbf{Image Editing.} The identified UPD factor area is then edited using Stable Diffusion, guided by a prompt template $P$. This template combines Target Words (\(Ta_W\)) (e.g., "Building", "Wall", "Vegetation") with Trigger Words (\(Tr_W\)) (e.g., "Beautiful") that describe desired improvements. (\(Ta_W\)) is selected from preset options, while (\(Tr_W\)) is iteratively tuned during the optimization loop.
    
    \item \textbf{Urban Renewal Evaluation.} The edited images are evaluated using three human perception models (safety, beauty, and liveliness) based on the Place Pulse 2.0 dataset \cite{dubey2016deep}. These models simulate human perception and provide feedback, quantifying the improvement of each urban renewal process.  

    \item \textbf{Urban Renewal Optimization.} The trigger words \(Tr_W\) are refined, and a new textual prompt $P$ is generated using Bayesian Optimization $B(P)$. This process continues until no further improvement in human perception scores is detected.

\end{itemize}

The overall framework is presented as follows:

\[
SVI_{\text{output}} = f_{\text{edit}}\left( B \left( f_{\text{w}} \left( P \right) \right) \right)
\]

where \( B(f_{\text{w}}(P)) \) applies Bayesian Optimization to generate a new prompt \( P \). \( f_{\text{edit}}(P) \) uses the new prompt \( P \) for image editing based on human perception evaluation. \( SVI_{\text{output}} \) is the final output image after the best editing is applied. For detection and feedback, the prompt \( P \) is defined as:

\[
P = 
\begin{cases} 
    \text{if } f_{u}(SVI_i) \text{ detects UPD:} & P = \text{Template}(Ta_W, Tr_W) \\
    \text{else:} & return
\end{cases}
\]

This pipeline drives urban imagery from disorder detection to human perception-aligned renewal, ensuring that urban spaces are revitalized to align with human preferences and well-being.

\subsection{Urban Renewal Quantitative Metrics: Assessing Human Perception within Street View Imagery}

Urban renewal is about physical transformation and the perceptual shift it brings about in the minds of the inhabitants and viewers. To quantify this perceptual shift, we use metrics that resonate with urban renewal objectives: safety, beauty, and liveliness.

Drawing from a precedent study \cite{zhang2018measuring}, we utilize machine learning to evaluate human perceptions associated with $SVI$, focusing on the selected metrics. By comparing these scores before and after renewal interventions, we can quantify the improvement or decline in human perceptions. This analysis offers urban planners and stakeholders a concrete measure of the effectiveness of their strategies, ensuring that interventions lead to genuine enhancements in both the physical and perceptual realms of urban spaces.




\subsection{Strategies for Urban Renewal Simulation: Text-Driven Image Editing}

Urban renewal, especially in enhancing street imagery, requires precise identification of areas for improvement and subsequent enhancement. This research employs two distinct strategies, as illustrated in Figure \ref{fig:strategies}:

\begin{figure}[!h]
 \centering
 \includegraphics[width=1\linewidth]{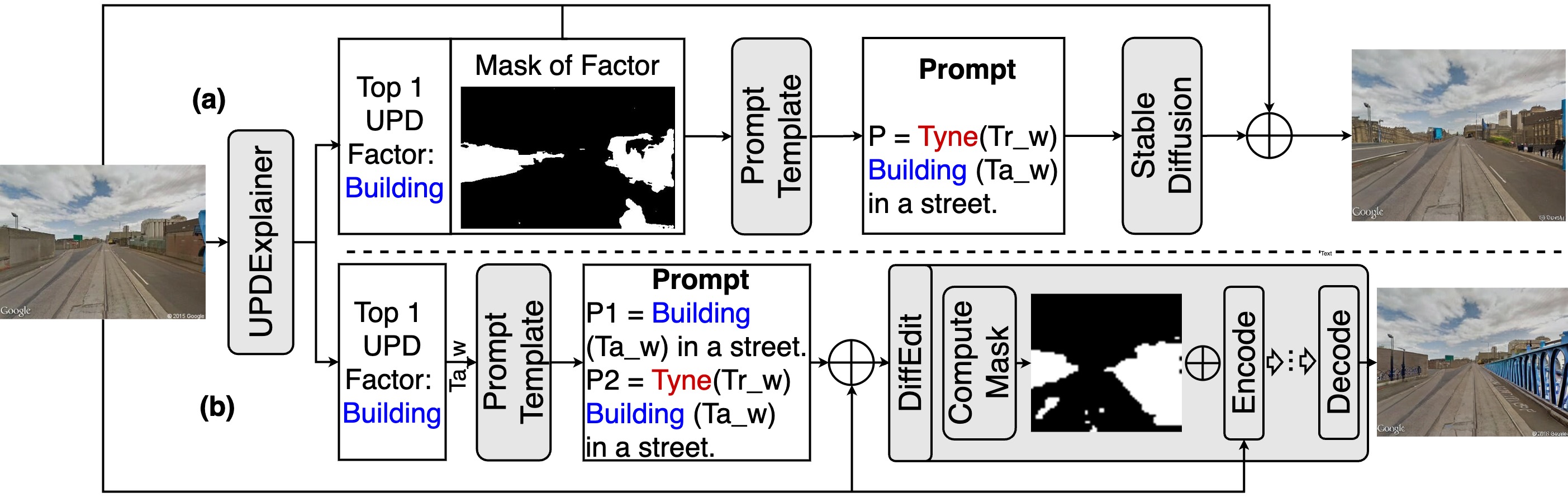}
 \caption{Two Strategies for Urban Renewal Simulation: Text-Driven Image Editing (a) Our strategy; (b) Diffedit}
  \label{fig:strategies}
\end{figure}

\begin{itemize}

\item \textbf{Our strategy (Figure \ref{fig:strategies} a)).} 

\begin{itemize}

\item UPDExplainer: Both strategies begin with the UPDExplainer module analyzing a street view image to detect the most prominent UPD factor. In this example, the identified factor is a "Building". After identifying the UPD factor, a mask is generated that specifically highlights the area of the image corresponding to this factor.

\item Prompt Generation: Using the identified UPD factor, a prompt is created by combining a trigger word (\(Tr_W\)) with a target word (\(Ta_W\)). For example, the prompt might be "Tyne Building in a street," where "Tyne" is the trigger word indicating the desired enhancement, and "Building" is the target word linked to the identified UPD factor.

\item Stable Diffusion: This prompt is then fed into a Stable Diffusion model, which uses the information to make targeted edits to the image, enhancing the appearance and improving the perceived quality of the identified building. The final output shows an improved version of the street environment.
\end{itemize}

 \item \textbf{DiffEddit \cite{couairon2022diffedit} (Figure \ref{fig:strategies} b)).} 

\begin{itemize}
    \item UPDExplainer: Similar to the first strategy, DiffEdit also begins by using UPDExplainer to detect the most significant UPD factor, which in this case is the component \(Ta_W\) ('Building') of the prompt.

    \item Text-Driven Mask Creation: Instead of using the factor to generate a precise mask, DiffEdit uses textual prompt $P1$ to automatically generate a mask. The prompt includes only \(Ta_W\)  ('Building'), which is used to guide the mask generation.

    \item Encoding and Decoding Process: DiffEdit processes the image by encoding the textual prompt, computing a mask from it, and then decoding the image with the mask applied based on the $P2$. The prompt includes \(Ta_W\) (e.g., "Building") and \(Tr_W\) (e.g., "Tyne"). This approach interprets the text and modifies the relevant areas of the image based on the overall prompt.

    \item Final Output: The image is edited according to the computed mask and the encoded text, resulting in an altered version of the street environment.
    
\end{itemize}

\end{itemize}

In both strategies, the success of urban renewal hinges on accurately identifying UPD factors through UPDExplainer and crafting precise prompts. These prompts include a target word \(Ta_W\) representing the UPD factor (e.g., "Building") and a trigger word \(Tr_W\) to guide the image edits. While the target word remains fixed, optimizing the trigger word is crucial for enhancing human perception metrics like safety and beauty. The next subsection will delve into how optimizing the trigger word can significantly improve the effectiveness and impact of our urban renewal simulations.

\subsection{Enhancing Urban Renewal: Prompt Tuning via Bayesian Optimization}

To maximize human perception enhancements in urban renewal simulations, optimizing the textual prompt $P$ is essential. The process of optimization refines the initial prompt by iteratively adjusting the trigger words \(Tr_W\), ultimately producing an optimal prompt \( P_{\text{optimal}} \). The optimal prompt \( P_{\text{optimal}} \) is defined as:

\[
P_{\text{optimal}} = \text{argmax}_{Tr_W} \, R \left( f_{\text{edit}}\left(\text{Template}(Ta_W, Tr_W)\right) \right)
\]

Where \( \text{Template}(Ta_W, Tr_W) \) is the prompt structure combining \( Ta_W \) and \( Tr_W \) into a cohesive textual instruction for image editing.  \( f_{\text{edit}}(P) \) is the function that applies the prompt \( P \) to the street view image \( SVI \), generating an edited image. \( R(\cdot) \) is the reward function that quantifies the improvement in human perception (e.g., safety, beauty, liveliness) after the image has been edited.

This equation expresses that the optimal prompt \( P_{\text{optimal}} \) is the one that maximizes the reward function \( R(\cdot) \), which is evaluated based on the human perception improvements resulting from the edited image.

Algorithm for Bayesian Optimization
The optimization process can be formalized through the algorithm \ref{alg:optimizeImageGeneration}. 

\begin{algorithm}[h]
\small
\caption{Bayesian Optimization for Urban Renewal Simulation}\label{alg:optimizeImageGeneration}
\begin{algorithmic}[1]
\Require Textual description $P$, number of iterations $n$
\Ensure Optimized prompt $P_{\text{optimal}}$ and final edited image $SVI_{\text{output}}$
\State Initialize $best\_score \gets -\infty$
\State Define the parameter space for tunable parameters
\State Define the objective function $\mathcal{O}(Tr_W)$:
    \State \hspace{0.4cm} Generate edited image $SVI'$ using $P$
    \State \hspace{0.4cm} Evaluate $SVI'$ using human perception metrics $M$
    \State \hspace{0.4cm} Compute reward $R(\Delta)$ based on change in metrics
    \State \hspace{0.4cm} \Return $-\mathcal{O}(Tr_W) = -R(\Delta)$ \Comment{Maximizing the reward}
\State Initialize Bayesian optimization framework with $\mathcal{O}(Tr_W)$
\For{iteration $i = 1$ to $n$}
    \State Select $Tr_W$ using Bayesian optimization to minimize $\mathcal{O}(Tr_W)$
    \State Evaluate $R(\Delta)$ and update $best\_score$ if improved
    \State Update the Bayesian model with the selected $Tr_W$ and corresponding $R(\Delta)$
\EndFor
\State Generate final image $SVI_{\text{output}}$ using $P_{\text{optimal}}$
\State \Return $P_{\text{optimal}}$, $SVI_{\text{output}}$
\end{algorithmic}
\end{algorithm}

By methodically refining the prompt through Bayesian Optimization, we ensure that the generated \( SVI_{\text{output}} \) not only addresses the visible Urban Physical Disorder but also aligns closely with human perceptions, enhancing the urban environment in terms of safety, beauty, and liveliness. This approach provides urban planners with a powerful tool to simulate and implement urban renewal strategies that are both effective and perceptually meaningful.

\section{Experiments}\label{sec:4}
\subsection{Experimental Settings}

This subsection outlines the setup used for conducting experiments, including details on the datasets, baselines, and hardware and software specifications.

\begin{itemize}
    \item \textbf{Datasets:}
Our experiments utilize a set of 500 SVI sourced from the Place Pulse 2.0 dataset. This dataset provides a diverse collection of urban scenes, ideal for evaluating our text-driven prompt tuning framework. Each image has been analyzed for urban physical disorder using the method described in \cite{hu2023updexplainer}. The dataset is divided into four distinct urban renewal scenarios: Neighborhood Improvement ($NI$), Building Redevelopment ($BR$), Green Space Expansion ($GSE$), and Community Gardens ($CG$). Each scenario is represented by 125 street-view images, allowing for a comprehensive assessment of our framework's effectiveness across various urban settings.

\item \textbf{Baselines:}
To evaluate the performance of our framework, we compared it against three baseline approaches:
\begin{itemize}
    \item Manually Designed Prompts ($MP$): Prompts created by human experts without the aid of AI serve as a standard for traditional text-driven image editing.
    \item Similar Words of Manually Designed Prompts ($SW$): This method utilizes prompts that are syntactically similar to the manually designed prompts, employing Word2Vec to generate variations and assess the robustness of our model against slight linguistic changes.
    \item DiffEdit \cite{couairon2022diffedit}: A method representing an alternative text-driven image editing approach. Unlike our framework, which uses image segmentation to extract masks before applying text-driven edits, DiffEdit directly generates and edits masks based on textual prompts. This comparison highlights the advantages of our pre-segmentation approach in enhancing the accuracy and relevance of text-driven modifications in urban scenes.
    
\end{itemize}

\item \textbf{Evaluation Metrics}
The performance of the proposed framework is evaluated based on improvements in three human perception metrics: safety, beauty, and liveliness. The improvement rate is calculated as:
\[
\text{Improvement Rate} = \frac{\text{Renewal Score} - \text{Previous Score}}{\text{Previous Score}}
\],
where the renewal score represents the human perception metric after applying the urban renewal interventions, and the previous score represents the metric before the interventions. This metric allows us to quantify the enhancement in perceived safety, beauty, and liveliness of urban environments.

\item \textbf{Hardware and Software:}

The experiments were conducted on a computing system equipped with dual NVIDIA GTX 3090 GPUs, providing the necessary processing power for image generation and manipulation tasks. We implemented our neural network models using the PyTorch framework, known for its flexibility and efficiency in handling large datasets and complex computations. Word2Vec was employed for generating and handling word embeddings, which was crucial in textual analysis and prompt generation.

\end{itemize}

\subsection{Comparison Analysis of Urban Renewal Strategies}
\subsubsection{Quantitative Evaluation of Urban Renewal Strategies}

This subsection presents a comparative analysis of our proposed framework against established baselines. The comparison focuses on improving human perception across three key dimensions: safety, beauty, and liveliness. Table \ref{tab:baseline} summarizes the performance of each model:

\begin{table}
\centering
\caption{Quantitative Comparison of the proposed strategies with baseline models for urban renewal. Bold font demonstrates the best performance}
\begin{tabular}{l|lll}
\hline
\multirow{2}{*}{Urban renewal baseline} & \multicolumn{3}{l}{Human perception improvement $\uparrow$} \\ \cline{2-4} 
                                        & Safe           & Beauty         & Lively        \\
\hline
$MP$                        & 4.05\%        & 15.29\%        & 19.67\%        \\
$SW$           
& 3.08\%            & 13.51\%            & 18.49\%           \\
DiffEdit                              & 2.31\%        & 11.87\%        & 15.84\%        \\
\hline
\textbf{Ours}                              & \textbf{17.60}\%        & \textbf{31.15}\%        & \textbf{28.82}\%        \\
\hline
\end{tabular}
\label{tab:baseline}
\end{table}

The results in Table \ref{tab:baseline} demonstrate that our framework significantly outperforms all baselines across all metrics. The substantial gains in safety, beauty, and liveliness underscore our framework's effectiveness in aligning virtual urban enhancements with human perceptual preferences, a critical factor in urban planning applications where visual and perceptual impacts are paramount.

\subsubsection{Qualitative Insights from Urban Renewal Case Studies}

To further illustrate the superiority of our framework over baseline models, we present detailed case studies in Table \ref{tab:case_baseline}. 

\begin{table}[h]
    \centering
    \small
    \begin{tabular}{c|c|c|c|c|c}
        \hline
        No. & Raw Image & $MP$ & $SW$ & DiffEdit & Ours \\
        \hline
        1   & \includegraphics[height=2.35cm,width=2.35cm]{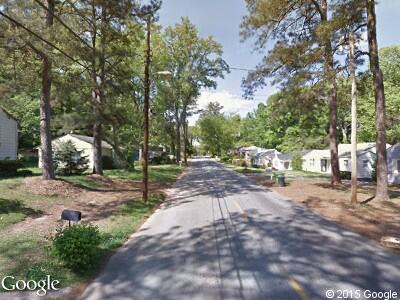} & \includegraphics[height=2.35cm]{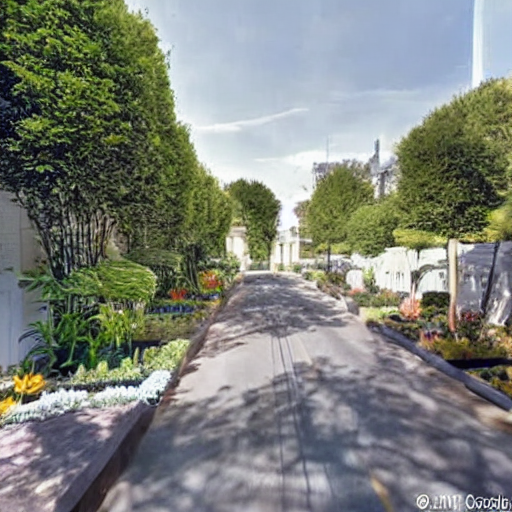} & \includegraphics[height=2.35cm]{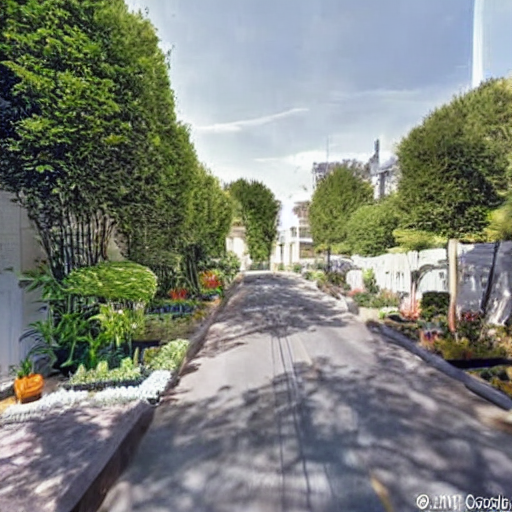} & \includegraphics[height=2.35cm]{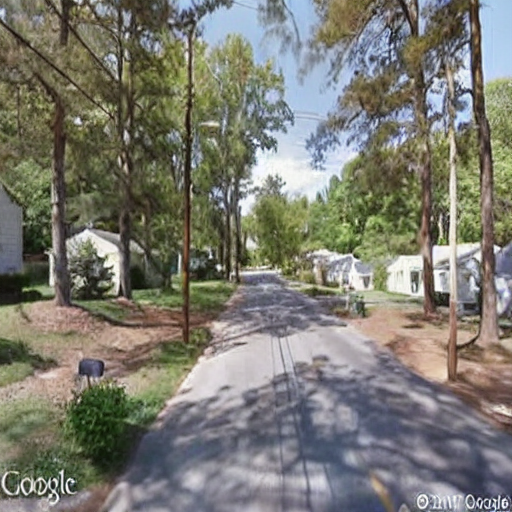} & \includegraphics[height=2.3cm]{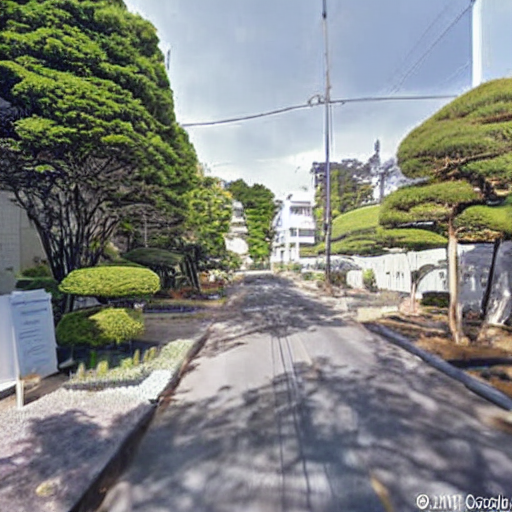} \\ \hline
        \(Tr_W\) & $/$ & Beautiful & Wonderful & Mayor & Mayor \\ \hline
        $S_{Beautiful}$ & 6.83 & 7.64  & 7.76 & 6.97 & 9.99 \\
        \hline
        2 & \includegraphics[height=2.35cm,width=2.35cm]{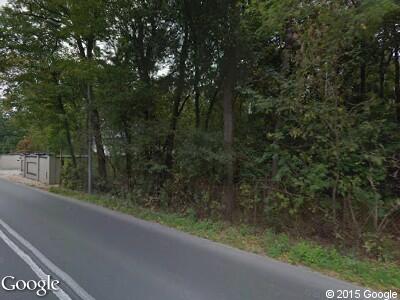} & \includegraphics[height=2.35cm]{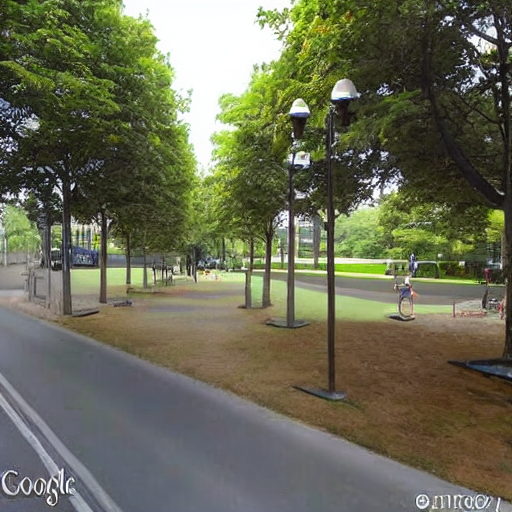} & \includegraphics[height=2.35cm]{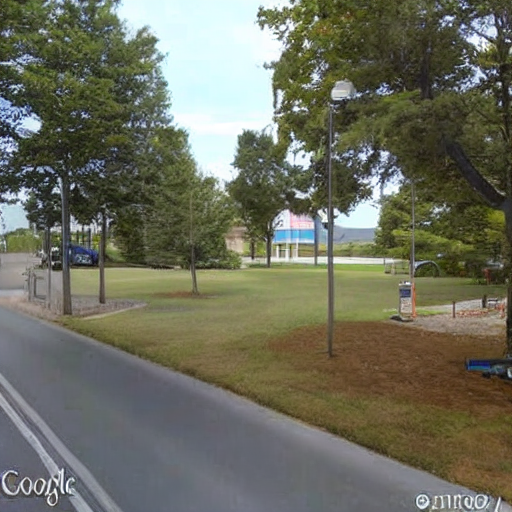} & \includegraphics[height=2.35cm]{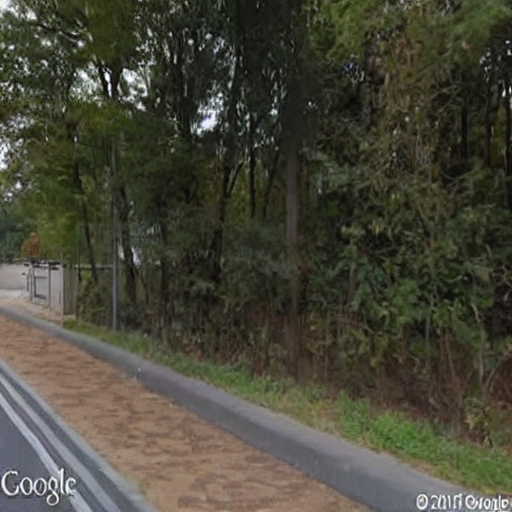} & \includegraphics[height=2.35cm]{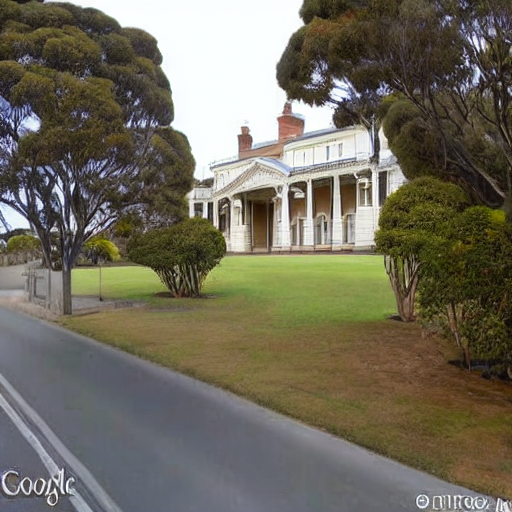} \\ \hline
        
        \(Tr_W\) & $/$ & Lively & Vacationland & Werribee &  Werribee  \\ \hline 
        $S_{Lively}$ & 4.37 & 4.51 & 4.83 & 4.05 & 6.79 \\
        \hline
        3 & \includegraphics[height=2.35cm,width=2.35cm]{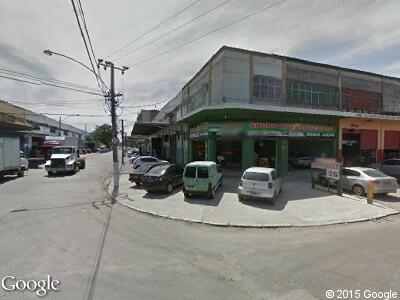} & \includegraphics[height=2.35cm]{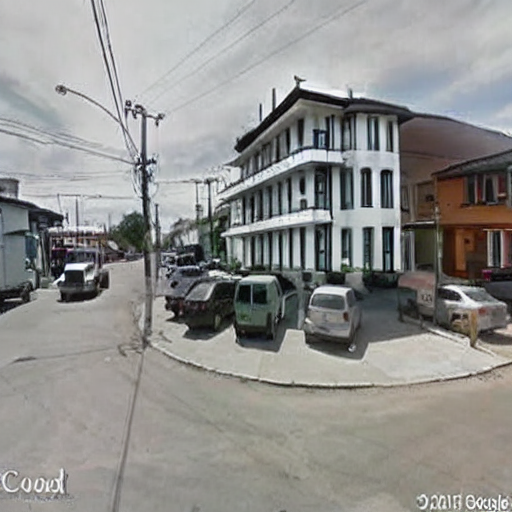} & \includegraphics[height=2.35cm]{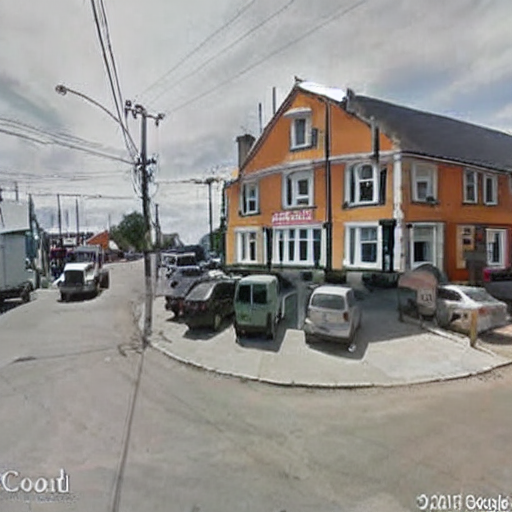} & \includegraphics[height=2.35cm]{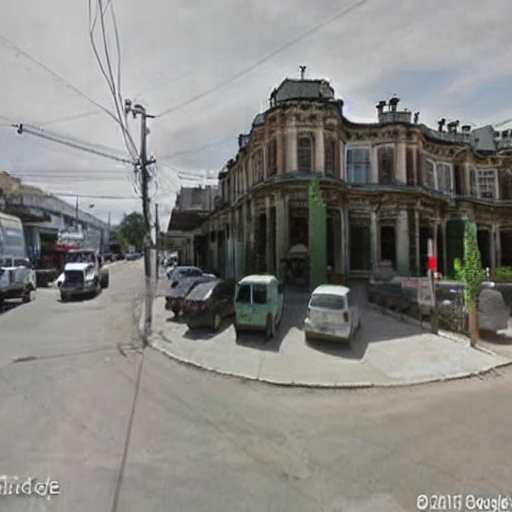} & \includegraphics[height=2.35cm]{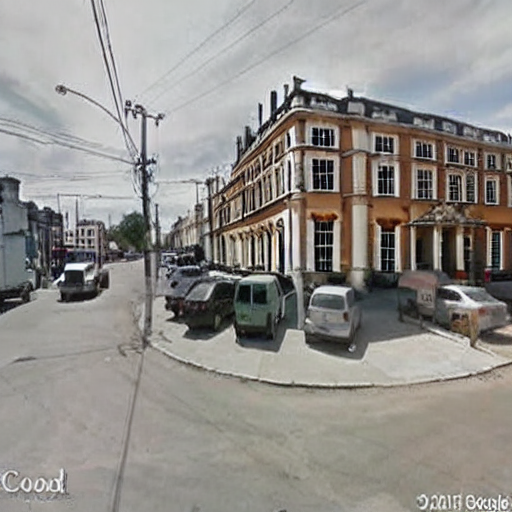}  \\ \hline
        \(Tr_W\) & $/$ & Safe & Coastguards & Gin Palaces & Gin Palaces \\ \hline
        $S_{Safe}$ & 4.48 & 5.00 & 5.31 & 4.54 & 5.35 \\ \hline
    \end{tabular}
    \caption{Performance Evaluation of Urban Renewal Strategies Across Visual Scenarios}
    \label{tab:case_baseline}
\end{table}

These studies provide visual and numerical comparisons, highlighting the improvements achieved by our approach over baseline models in real-world urban scenarios. Observations based on Table \ref{tab:case_baseline} are as follows:

\begin{itemize}
    \item \textbf{Case 1:} 
    Our framework significantly enhances the visual beauty of the community garden scenario, achieving the highest beauty score of 9.99, substantially outperforming other methods. This demonstrates the framework's ability to elevate urban green spaces' aesthetic appeal.

   \item \textbf{Case 2:}
   our method improved the liveliness score from 4.37 to 6.79, showcasing the effectiveness of our text-driven modifications in creating vibrant, engaging urban spaces.

   \item \textbf{Case 3:}
   Our framework successfully enhances perceived safety on a downtown street corner through subtle visual modifications, highlighting its potential in improving urban environments' safety perceptions.
   
\end{itemize}

These case studies underscore the practical impact of our text-driven prompt tuning framework, particularly in its ability to resonate with human perception. The observed improvements are quantitatively significant and visually apparent, making a compelling case for the framework's application in real-world urban planning.

\subsection{Comparison Analysis of Different Scenarios}

\subsubsection{Quantitative Evaluation based on Different Scenarios}

This subsection evaluates the effectiveness of our urban renewal strategies across four distinct scenarios: Neighborhood Improvement ($NI$), Building Redevelopment ($BR$), Green Space Expansion ($GSE$), and Community Gardens ($CG$). Each scenario focuses on specific changes to urban elements and is assessed based on improvements in human perception metrics: safety, beauty, and liveliness.

Table \ref{tab:segment} presents a comparative evaluation of how our urban renewal strategies enhance human perceptions of safety, beauty, and liveliness across these scenarios. Each scenario transforms specific urban features, demonstrating targeted improvements in the urban landscape.

\begin{table}[]
\small
\centering
\caption{Evaluation of Different Scenarios for Urban Renewal}
\begin{tabular}{l|l|lll}
\hline
\multirow{2}{*}{\begin{tabular}[c]{@{}l@{}} $UR$ \\ Scenario\end{tabular}} & \multirow{2}{*}{\begin{tabular}[c]{@{}l@{}}Semantic object\\ change\end{tabular}} & \multicolumn{3}{l}{Human perception + $\uparrow$} \\ \cline{3-5} 
                                        &                                                                                   & Safety             & Beauty             & lively           \\ \hline
$NI$             & Wall/Fence -\textgreater Wall/Fence                                                & 13.16\%            & 15.06\%            & 8.27\%           \\
$BR$                 & Building -\textgreater Building                                                   & 15.83\%            & 26.47\%            & 18.80\%           \\
$GSE$                 & Vegetation -\textgreater Park                                                     & 19.79\%            & 44.66\%            & 46.10\%           \\
$CG$                     & Vegetation -\textgreater Gardens                                                     & 16.16\%            & 37.34\%            & 40.81\%           \\ \hline
\end{tabular}
\label{tab:segment}
\end{table}

\begin{itemize}
    \item $NI$: Focuses on enhancing perimeter security and aesthetic features by transforming walls and fences, which has shown modest improvements across all three metrics, particularly in safety and beauty.
    
    \item $BR$: Targets structural facades and functionalities, leading to significant gains in beauty and liveliness, reflecting the high impact of architectural aesthetics on urban perception.
    
    \item $GSE$: By converting unused or underutilized vegetation into parks, this scenario achieves the greatest improvements, particularly in making urban areas feel more lively and attractive. The transformation is reflected in dramatically higher scores for both beauty and liveliness.
    
    \item $CG$: Like green space expansion, converting general vegetation into organized community gardens significantly enhances perceptions of beauty and liveliness, fostering community engagement and environmental quality.
\end{itemize}

These results highlight the tailored effectiveness of different urban renewal strategies in improving the human experience in urban environments. By addressing specific urban elements, these interventions enhance the visual and functional aspects of cityscapes and elevate the overall quality of life for residents.

\subsubsection{Qualitative Insights based on Different Scenarios}

This subsection offers qualitative insights from specific urban renewal scenarios, focusing on the perceptual changes before and after the renewal interventions. The comparisons across the four distinct scenarios — $NI$, $BR$, $GSE$, and $CG$ — are visually and numerically detailed in Table \ref{tab:case_scenario}.

\begin{table}[h]
    \centering
    \small
    \begin{tabular}{c|c|c}
        \hline
        Scenario & Raw Image & Renewal Image \\
        \hline
        $NI$ & \includegraphics[height=1.6cm]{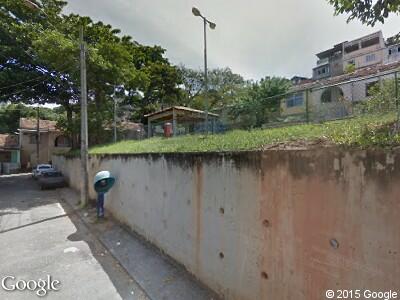} & \includegraphics[height=1.6cm,width=2.2cm]{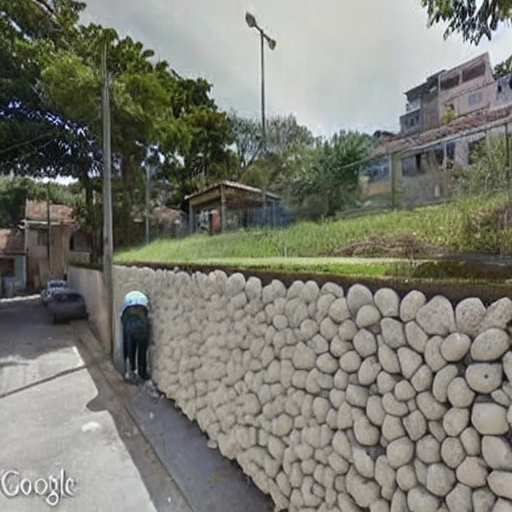}  \\ \hline
        $S_{Safe}$ & 4.78 & 5.97 \\ \hline 

        $BR$ & \includegraphics[height=1.6cm]{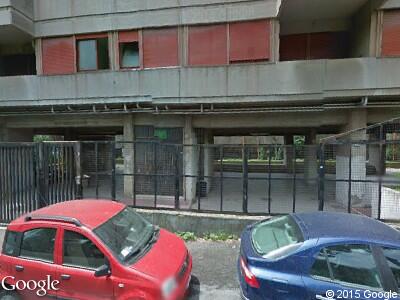} & \includegraphics[height=1.6cm,width=2.2cm]{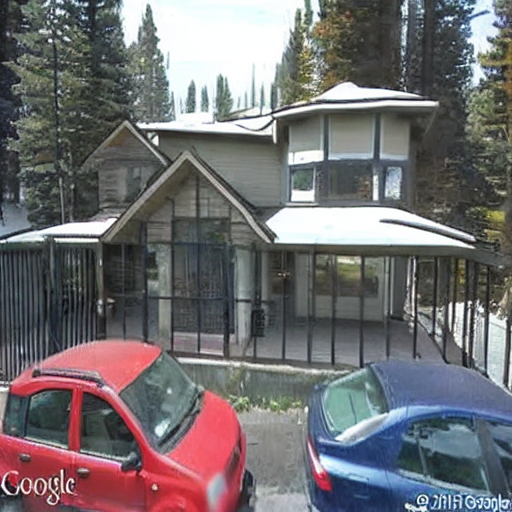}   \\ \hline
        $S_{Lively}$ & 5.59 & 7.65\\ \hline

        $GSE$ & \includegraphics[height=1.6cm]{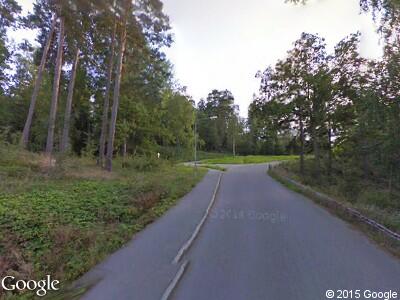} & \includegraphics[height=1.6cm,width=2.2cm]{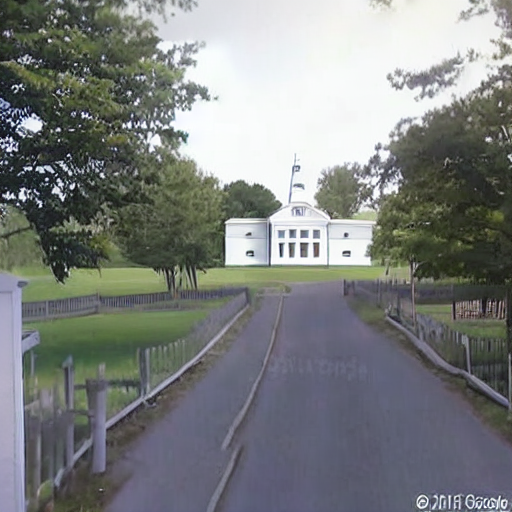}  \\ \hline
        $S_{Beautiful}$ & 6.08 & 7.91 \\ \hline
        $CG$ & \includegraphics[height=1.6cm]{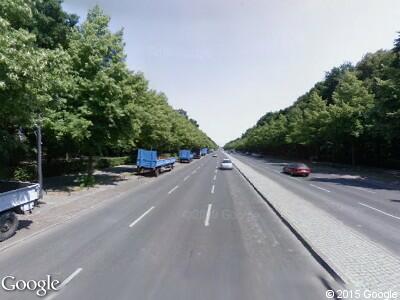} & \includegraphics[height=1.6cm,width=2.2cm]{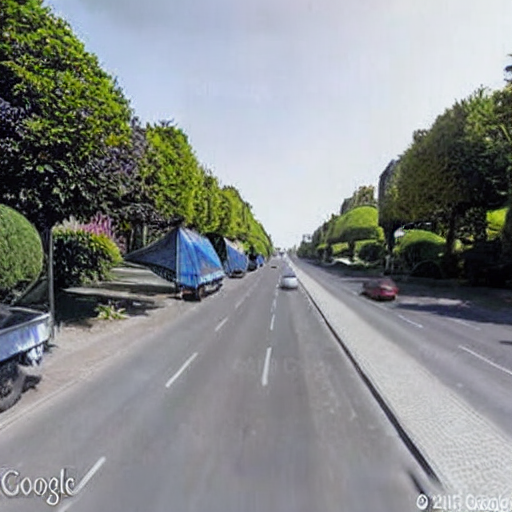}  \\ \hline
        $S_{Beautiful}$ & 7.79 & 8.93 \\ \hline

    \end{tabular}
    \caption{Visual and Numerical Comparison Across Scenarios}
    \label{tab:case_scenario}
\end{table}

Table \ref{tab:case_scenario} provides a visual depiction and numerical scores illustrating the improvements in perceptions of safety, liveliness, and beauty following urban renewal interventions. The scores reflect the community's or evaluators' perceived changes, offering tangible evidence of the improvements each renewal scenario brings.

\begin{itemize}
    \item $NI$: Focusing on safety, the renewal involved replacing a deteriorated wall with a more secure and aesthetically pleasing fence. This intervention increased the safety score from 4.78 to 5.97, indicating a more secure environment perceived by residents.
    
    \item $BR$: Targeting liveliness, the intervention transformed a standard urban building facade into a more vibrant and inviting structure. The liveliness score improved from 5.59 to 7.65, reflecting a more dynamic and engaging urban space.
    
    \item $GSE$: This scenario aimed at enhancing beauty by converting an underutilized vegetated area into a well-maintained park. The beauty score rose from 6.08 to 7.91, showcasing the aesthetic enhancement and the attractiveness of the space.
    
    \item $CG$: Similar to GSE, this renewal turned an ordinary vegetated area into a community garden, boosting the beauty score from 7.79 to 8.93. This beautified the area and promoted community engagement and environmental stewardship.

\end{itemize}

These case studies demonstrate the impactful role of specific urban renewal interventions in transforming urban environments. The visual comparisons underline the effectiveness of thoughtful design and targeted improvements in enhancing urban spaces' perceived safety, liveliness, and beauty. Such interventions not only elevate the aesthetic and functional quality of these areas but also contribute significantly to the well-being and satisfaction of the community. These qualitative insights, supported by numerical data, offer compelling evidence of the benefits of our urban renewal strategies, reinforcing the value of our approach in real-world applications.

\subsection{Evaluation of Sensitivity to Urban Morphology}
\subsubsection{Quantitative Evaluation Based on Urban Morphology}

This subsection explores the influence of urban morphology, quantified through the height-to-width ratio (H/W, denoted as $\alpha$), on the effectiveness of urban renewal strategies. Street view images can be classified into different H/W categories \cite{hu2020classification}, which facilitates comparisons in this section. The evaluation is conducted through quantitative metrics and visual comparisons to assess the impact on human perceptions of safety, beauty, and liveliness.

Table \ref{tab:quantitative_h/w} provides a quantitative evaluation of the influence of the H/W ratio on the performance of urban renewal strategies across different urban scenarios.

\begin{table}[]
\centering
\caption{Quantitative evaluation of the influence of $\alpha$ on the performance for urban renewal.}
\begin{tabular}{l|lll}
\hline
\multirow{2}{*}{Urban Morphology} & \multicolumn{3}{l}{Human perception improvement $\uparrow$} \\ \cline{2-4} 
                                        & Safety             & Beauty             & lively           \\ \hline
Barely Populated ($\alpha < 0.5$)                              & 18.13\%            & 40.61\%            & 42.34\%           \\
Living Spaces ($0.5 < \alpha \leq1.5$)                        & 14.55\%            & 22.76\%            & 19.10\%           \\
Urban Hub ($\alpha>1.5$)                        & 15.16\%            & 32.49\%            & 27.37\%           \\ \hline
\end{tabular}
\label{tab:quantitative_h/w}
\end{table}

Table \ref{tab:quantitative_h/w} indicates that suburban areas with a lower H/W ratio, classified as Barely Populated, see the greatest improvement in all human perception metrics post-renewal. This aligns with real-world observations, where enhancing human perception in older urban areas through green space expansion or community gardens is challenging. In contrast, suburban areas respond more positively to urban renewal efforts. Additionally, the proposed method performs better in high-density urban areas than with common density. To highlight the varying performance of the proposed method across different urban morphologies, Table \ref{tab:case_h/w} presents several cases, further illuminating how changes in urban morphology influence renewal outcomes.

\begin{table}[h]
    \centering
    \small
    \caption{Visual comparison of urban renewal impacts across varying urban morphologies.}
    \begin{tabular}{c|c|c}
        \hline
        $H/W$ & Raw Image & Renewal Image  \\
        \hline
        $\alpha < 0.5$ & \includegraphics[height=3cm]{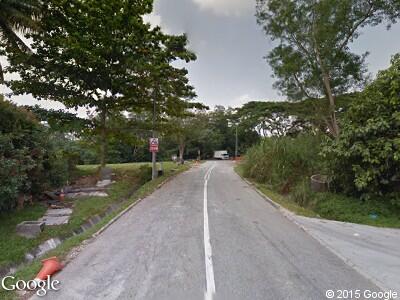} & \includegraphics[height=3cm,width=4.2cm]{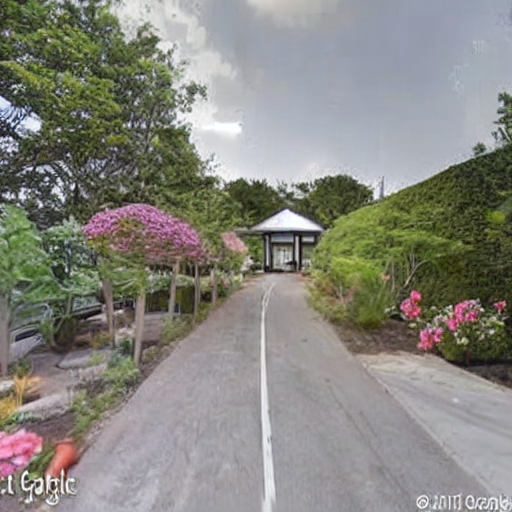} \\ \hline Beautiful & 6.32 & 8.94 \\ \hline
        $0.5 < \alpha \leq1.5$ & \includegraphics[height=3cm]{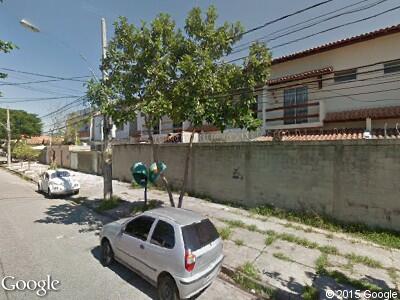} & \includegraphics[height=3cm,width=4.2cm]{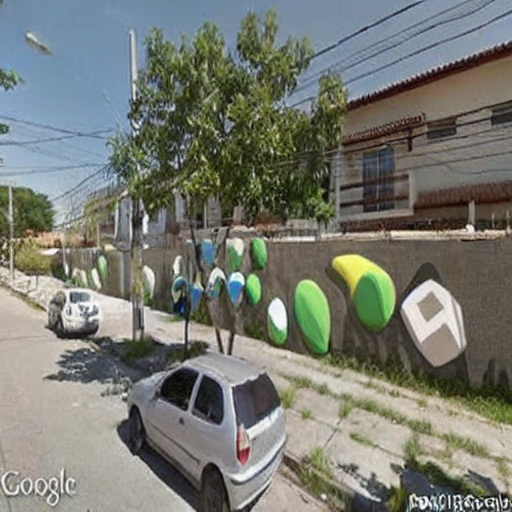} \\ \hline
        Safe & 4.05 & 4.91 \\ \hline
        $\alpha>1.5$ & \includegraphics[height=3cm]{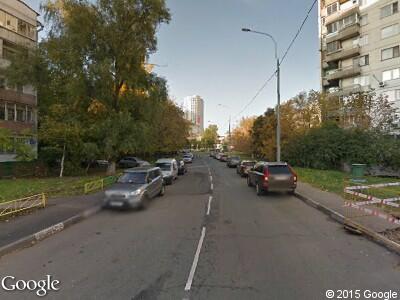} & \includegraphics[height=3cm,width=4.2cm]{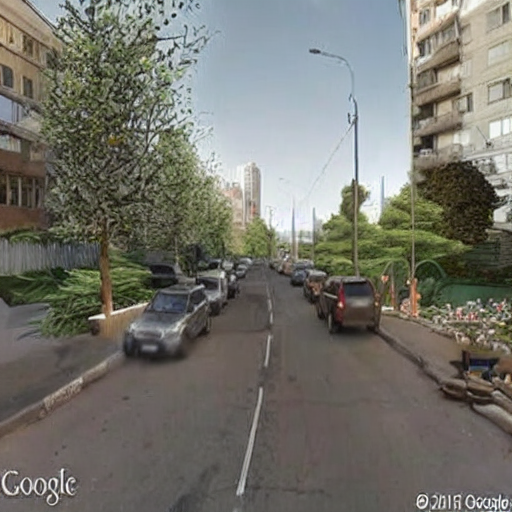} \\ \hline
        Lively & 5.17 & 6.37  \\ \hline
    \end{tabular}
    
    \label{tab:case_h/w}
\end{table}

The visual comparisons provide a clear before-and-after snapshot, highlighting the enhancements in aesthetics and functionality. Each scenario, influenced by its respective H/W ratio, shows a unique set of improvements, with scenarios having a lower $\alpha$ generally exhibiting more pronounced enhancements. This visual evidence supports the quantitative findings and further emphasizes the role of urban morphology in determining the success of urban renewal efforts.

\section{Discussions}\label{sec:5}  

The proposed framework for text-driven image editing offers significant advancements in urban renewal. This section discusses the key findings from our experiments, their broader implications, the limitations of our research, and potential directions for future work.

\subsection{Enhanced Accuracy Through Object-Level Interventions}

Our approach, which involves extracting and segmenting specific semantic objects before performing image editing, outperforms direct text-driven methods like DiffEdit \cite{couairon2022diffedit}. Our method ensures more precise and relevant modifications by focusing on targeted objects within the urban environment that require enhancement. This pre-segmentation step is crucial, leading to higher improvements in human perception metrics. The success of this approach underscores the importance of object-level interventions in urban renewal, demonstrating a clear advantage over methods that skip this step and apply edits directly based on text prompts.

\subsection{Optimizing Textual Prompts for Enhanced Perception}

One of the key insights from our work is the gap between human-intuitive text-driven words and those optimized for diffusion models. While words like "beautiful" may seem effective from a human perspective, our experiments show that diffusion models often respond better to less obvious words such as "mayor" (see Table \ref{tab:case_baseline}). This highlights the importance of optimizing textual prompts to achieve the best outcomes in human perception scores. Our method’s ability to identify and utilize these optimal words is critical to its effectiveness, suggesting that human intuition alone may not align with computational optimization. This finding reinforces the value of our prompt optimization process, which significantly enhances task performance over human-selected prompts, as supported by prior research \cite{radford2021learning}.

\subsection{Significance difference of Scenarios in Urban Renewal}

Our analysis reveals that urban renewal efforts focused on vegetation significantly improve human perception metrics compared to interventions targeting urban facilities like old buildings, fences, and walls. This finding aligns with existing research emphasizing the benefits of green space expansion and reconstruction. Vegetation enhances aesthetic appeal and positively impacts perceived safety and liveliness, making it a highly effective component of urban renewal strategies. Previous studies have demonstrated green spaces' psychological and social benefits in urban environments, including reduced crime rates and improved mental health \cite{kuo2001environment,larkin2019evaluating}.

\subsection{Impact Variations Across Urban Morphologies}

Our evaluation across different urban morphologies shows notable variations in the effectiveness of our framework. Specifically, barely populated areas exhibit the highest improvements in all human perception metrics following renewal interventions. This observation aligns with real-world trends where suburban areas typically respond more positively to urban renewal efforts than densely populated urban areas. Additionally, our method performs better in high-density urban environments than in moderate-density ones. These insights suggest that our framework is particularly effective in transforming high-density and suburban environments, offering valuable guidance for urban planners to tailor renewal strategies according to the specific characteristics of the urban area. Similar findings have been reported in studies assessing the impact of urban renewal across different urban forms \cite{donovan2012effect,wolf2016benefits}.

\subsection{Limitations and Future Directions}

Despite the promising results, several limitations to our research also point toward future directions for improvement:

\begin{itemize}
\item Dataset Size and Diversity: Our experiments were conducted on a relatively small dataset of 500 street view images. While these images cover diverse urban scenarios, the sample size is limited. Future research should incorporate larger and more diverse datasets to validate the framework's robustness and generalizability. Expanding the dataset will help understand the framework's performance across various urban environments.

\item Textual Prompt Complexity: Although our method optimizes textual prompts effectively, designing more complex and varied prompts remains challenging. Further research is needed to explore automated methods for generating more sophisticated and context-aware prompts. Developing algorithms that dynamically generate and refine prompts based on evolving urban conditions will enhance the framework's adaptability and effectiveness.

\item Multi-Modal Data Integration: Combining visual data with other data sources, such as socio-economic and environmental data, could provide a more comprehensive approach to urban renewal. This integration would enable more informed decision-making and planning, considering a wider array of factors that influence urban environments.

\end{itemize}

\section{Conclusions}\label{sec:6}
This study presents a novel framework for urban renewal that leverages text-driven image editing techniques to enhance human perceptions of urban environments. By extracting semantic objects from street view images and optimizing textual prompts for image editing, our approach significantly improves over traditional methods such as DiffEdit. Furthermore, the research underscores the importance of optimizing textual prompts, revealing a notable gap between human-recognized effective words and those optimized for diffusion models. This optimization is crucial for achieving substantial improvements in human perception scores. Additionally, the study demonstrates that urban renewal efforts focused on vegetation yield more significant improvements than interventions targeting urban facilities, aligning with existing research on the benefits of green space expansion. The framework also shows varying effectiveness across different urban morphologies, with barely populated and high-density areas responding more positively to renewal efforts.

In summary, the proposed framework offers a robust and flexible tool for urban renewal, significantly enhancing human perceptions of urban spaces. Integrating advanced image editing techniques with optimized textual prompts provides a promising direction for future urban renewal projects, aiming to create more livable, attractive, and sustainable urban environments. The demonstrated effectiveness of this approach underscores its potential to contribute meaningfully to urban planning and development. By enhancing human perceptions of safety, beauty, and liveliness, the framework aims to improve the quality of life for city residents through thoughtful and targeted urban renewal initiatives. As urban planners and policymakers seek to revitalize urban areas, this framework offers valuable insights and a practical tool for achieving these goals.

\section*{ACKNOWLEDGEMENTS}\label{ACKNOWLEDGEMENTS}

The NSF partially supports this work through grants CMMI-2146015 and CCSS-2348046 and the WV Higher Education Policy Commission Grant (HEPC.dsr.23.7).

 \bibliographystyle{elsarticle-num} 
 \bibliography{cas-refs}





\end{document}